\documentclass[letterpaper, 10 pt, conference]{ieeeconf}  

\IEEEoverridecommandlockouts                              

\usepackage{graphics} 
\usepackage{epsfig} 
\usepackage{amsmath} 
\usepackage{amssymb}  
\usepackage{algorithm}
\usepackage{verbatim}
\usepackage[noend]{algpseudocode}
\usepackage{subfig}
\title{\LARGE \bf Laser2Vec: Similarity-based Retrieval for Robotic Perception Data}

\author{Samer B. Nashed$^{1}$
\thanks{$^{1}$Samer Nashed is with the College of Information and Computer Sciences,
        University of Massachusetts, Amherst, MA 01003, USA.
        Email: {\tt\small snashed@cs.umass.edu}}}%
        
\begin{document}

\bibliographystyle{abbrv}

\maketitle
\thispagestyle{empty}
\pagestyle{empty}

\begin{abstract}
As mobile robot capabilities improve and deployment times increase, tools to analyze the growing volume of data are becoming necessary. Current state-of-the-art logging, playback, and exploration systems are insufficient for practitioners seeking to discover systemic points of failure in robotic systems. This paper presents a suite of algorithms for similarity-based queries of robotic perception data and implements a system for storing 2D LiDAR data from many deployments cheaply and evaluating top-k queries for complete or partial scans efficiently. We generate compressed representations of laser scans via a convolutional variational autoencoder and store them in a database, where a light-weight dense network for distance function approximation is run at query time. Our query evaluator leverages the local continuity of the embedding space to generate evaluation orders that, in expectation, dominate full linear scans of the database. The accuracy, robustness, scalability, and efficiency of our system is tested on real-world data gathered from dozens of deployments and synthetic data generated by corrupting real data. We find our system accurately and efficiently identifies similar scans across a number of episodes where the robot encountered the same location, or similar indoor structures or objects.


%
\end{abstract}

\vspace{-2mm}

\section{Introduction}

\vspace{-1mm}

Modern robots are capable of extended deployments \cite{biswas20161}, during which they may gather large amounts of data. Often, this data is either thrown out after becoming obsolete with respect to the current task, or is stored in log files \cite{thomas2014rosbag}. This data is crucial for discovering systemic failures in robot behavior, but current log technology is not conducive to such discoveries. Logs must be played back in (roughly) real-time, making analyzing weeks or months worth of deployment data infeasible. Furthermore, it may be easy for a human to tell qualitatively what is causing failures, but difficult to write precise definitions which flag the correct data. For instance, by playing a single log, a practitioner might see a robot getting lost at a particular hallway intersection when a human obstructs a salient feature. However, writing a script to process all previous data and check for this scenario is challenging if not impossible. Currently, the only way for a practitioner to check if such a failure is systemic is to watch every log file, or to write such a script.

Ideally, raw sensor data from a robot, with or without meta data, could be compressed and stored in a database such that fast, accurate, similarity-based queries could be used to explore log data for events indicative of systemic failures. This task can be broken down into three components. First, raw sensor data can be compressed while preserving information useful for future queries. Second, computing similarity between data instances requires a distance or similarity function. And third, depending on the algorithms used for compression and similarity calculation, returning high-quality top-k results in sub-linear time may require a non-trivial query evaluator. This paper proposes a suite of algorithms for these tasks and demonstrate their efficacy on a subset of robotic perception data, namely 2D LiDAR data.

\begin{figure}[t]
\includegraphics[width=\columnwidth]{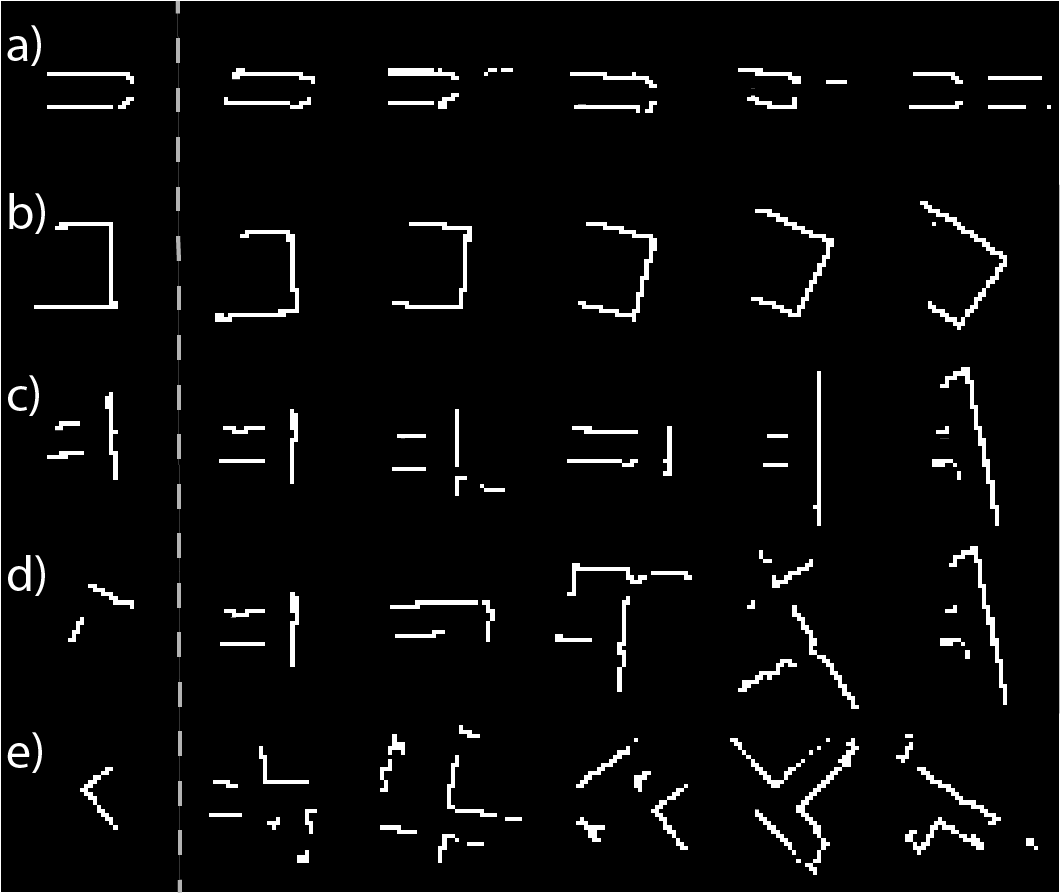}
\vspace{-5mm}
\caption{Laser returns (white) in robot frame. Images are downsampled 4x for display purposes. Queries (left) for scans of a) hallway with fire door, b) rectangular room, c) T-junction, and partial scans of d) a right turn and e) a large convex corner. A subset of the top-k results follow from left to right. Note that T-junctions contain right turns, and we find common items in the top-k for queries c) and d).}
\vspace{-7mm}
\label{fig:teaser}
\end{figure}

Neural networks are both effective data compression tools as well as distance function approximators. Much like word2vec \cite{mikolov2013distributed}, we learn embedded representations of laser scans via neural networks, which are then stored in a database. A variational autoencoder is used to generate the embeddings, and a dense network is trained as a siamese network to compute similarity between embeddings at query time. Because the learned distance function is non-metric, we cannot employ query evaluation based on exact tree pruning. Instead, we use the empirically supported local continuity of the embedded space to opportunistically search the neighborhood of high-scoring tuples found during evaluation.

Our system answers top-k queries where, given a query scan $q$, the $k$ most similar scans in the database are returned. The query $q$ may also be a contiguous subset of a scan, selected by the user. We provide both qualitative results, see Figure \ref{fig:teaser}, and quantitative experiments designed to test the accuracy, robustness, scalability, and efficiency of the system. 

\section{Related Work}

\vspace{-3mm}
Various architectures have been proposed for generating embeddings, many of which are derivatives of the autoencoder \cite{hinton2006reducing} or convolutional networks such as AlexNet \cite{krizhevsky2012imagenet}. One popular regularization method in autoencoders is the variational autoencoder (VAE) \cite{kingma2013auto}. In this paper, we use a convolutional variational autoencoder. Operating in embedding space has become an attractive way to deal with high-dimensional input, especially when similarity functions are difficult to define on the input space. Robotic sensors such as cameras and lasers have these traits, motivating an array of methods for visual retrieval \cite{babenko2014neural,razavian2016visual}, and recognition \cite{azizpour2015generic,wilkinson2015efficient,jang2018grasp2vec}. Embedded representations for depth sensors in particular have been used to compute motor commands \cite{pfeiffer2017perception}, estimate odometry \cite{nicolai2016deep}, predict loop closure \cite{li2017deep}, and even predict the presence of glass \cite{jiang2017glass}. The goal of Laser2Vec is not to outperform a network or embedding designed for a specific task. Rather, it is to encode representations of laser scans which support queries about the general similarity of scans or subsets of scans in support of data exploration.

Any robot deployed for an extended period of time will generate more data than may be stored and queried easily on the robot \cite{nelson2016building}. Thus, most robotic systems of the future will maintain associated databases storing past sensor data. The nascent interdisciplinary work between robotics and databases includes knowledge engines \cite{saxena2014robobrain}, failure provenance \cite{niemueller2012generic}, and notably, Vroom \cite{moll2017exploring}, a project with similar ambitions. To handle robotic perception data, Vroom memoizes the incoming data using pre-existing classifiers, typically done onboard the robot. This has the advantage of vastly compressing the data, as well as transforming it into a format suitable for relational database systems. However, it assumes that A) pre-existing classifiers are accurate, which may or may not be true, and B) that all future questions can be answered by queries expressible in terms of the existing classes, which is almost certainly false. Our approach therefore attempts to learn a representation for raw perception data which can more flexibly support future queries.

One problem introduced by learning distance functions over embedded representations is that the distance functions are not guaranteed to be metric. The lack of a triangle inequality makes efficient query evaluation challenging. There has been substantial research on top-k queries \cite{ilyas2008survey}, but the particular problem of real-valued non-metric search has not been researched as thoroughly. Related efforts include strategies for searching real-valued vectors under \emph{normed} distances \cite{zhu2016evaluating}, increasing index efficiency in high-dimensional space by using subspace indexes that can span multiple dimensions \cite{yu2016top}, 
and strategies for computing top-k queries under arbitrary compositions of arbitrary similarity measures \cite{lange2011efficient}. For tuples with categorical attributes, which may be the case for some meta-data generated by robot perception systems, the Attribute Level tree evaluates top-k \cite{deshpande2008efficient} and RkNN \cite{deshpande2010efficient} queries efficiently. Other approaches, such as \cite{xin2007progressive}, consider the best sub-trees in an indexing scheme to expand online, similar in spirit to the proposed method.

\vspace{-2mm}
\section{Laser2Vec Overview}

The Laser2Vec system is composed of two components: a preprocessor, and a query evaluator. Data from a robot, in this case 2D LiDAR data, is preprocessed before being stored in a database, where it is accessed at runtime by the query evaluator. The preprocessing step first converts the depth scan into a bitmap representing the discretized cartesian coordinates of all laser returns and then passes the bitmap through a convolutional variational autoencoder. The intermediate representation of the laser scan bitmap is extracted and stored in the database. At query time, a query vector $q$ is created from query scan $s_q$ via the same network, and the query evaluator computes the similarity between $q$ and all scans $s_i$ in the database using a separate network. The order in which scans are compared (the selection of $i$ at each step) is governed by a greedy Monte Carlo algorithm which adaptively samples parts of a sparse multi-graph representing the data items. The $k$ most similar vectors are returned as the result of the query. In the following sections, we present the autoencoder used for compression, the network used to compute similarity, and the query evaluator in detail. 


\section{LIDAR Compression}
\vspace{-2mm}
\subsection{Architecture}

We found that even deep, fully connected networks, and networks similar to 1D versions of inceptionNet \cite{szegedy2015going} have difficulty learning consistent internal representations when given raw scans in polar coordinates, possibly since most features from indoor laser scans are rectilinear and thus highly non-linear in polar coordinates. Creating a bitmap of registered observations transforms scans to cartesian coordinates and allows the use of 2D convolutional networks, which have been shown to learn useful internal representations for a variety of tasks.

\begin{figure}
\includegraphics[width=\columnwidth]{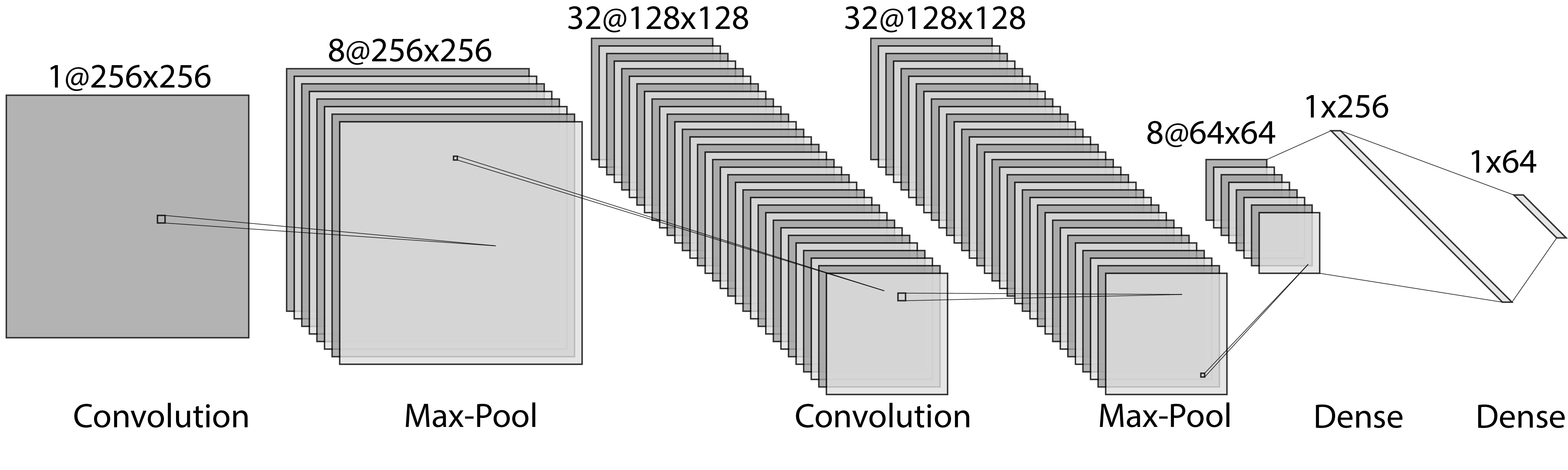}
\vspace{-8mm}
\caption{Variational autoencoder architecture used to encode laser scans. Four convolutional layers are followed by two fully connected layers, resulting in mean and standard deviation vectors in $\mathbb{R}^{32}$.}
\vspace{-8mm}
\label{fig:vae}
\end{figure}

The architecture of our compression network is shown in Fig \ref{fig:vae}. Four convolution layers with ReLU activation functions followed by two fully connected layers are applied in sequence to a 256x256 bitmap, resulting in an embedded vector in $\mathbb{R}^{32}$. The decoder portion of the network is symmetric, with two fully connected layers followed by four deconvolution layers. A final HardTanh function is applied to the reconstructed output prior to computing the loss.

\subsection{Training}

The network is trained on roughly 50K laser scans, with 20$\%$ held out for validation, and 20$\%$ held out for testing. To increase the network's ability to generalize, rotations, flips and subset selections are applied to the inputs and reconstruction targets as they are loaded at training time. Subset selections artificially restrict the laser's field of view to a random, contiguous subset of its original field of view.

Training loss is computed as the sum of the recreation loss and the KL-divergence between the latent vector and the standard Gaussian. Because most locations do not contain a laser return, the pixel-level classification in the reconstruction is unbalanced. To address this, the recreation loss weights failing to recreate an observation more heavily than erroneously labeling an empty pixel as containing an object. We find that weighting based on the proportion of observations to non-observations produces poor recreations, and instead achieve better performance using a weight ratio of 50:1. Inputs are fed to the network in shuffled batches of 128, and weight updates are computed using Adam \cite{kingma2014adam}. 
\vspace{-1mm}


\section{Similarity Learning}
\vspace{-2mm}
\subsection{Architecture}

To compute the similarity between two embedding vectors, we use a second neural network. We find that learning a distance function in embedding space is easier, more robust, and more accurate than training the embedding vectors directly and using a simpler distance function such as Euclidean distance or cosine distance. The use of a function approximator instead of a closed-form distance metric allows us to train the compression procedure and the similarity calculation independently, which makes retraining on updated data faster and ablation testing and parameter searches simpler, due to the de-coupling of most hyper-parameters.

\begin{figure}
\includegraphics[width=\columnwidth]{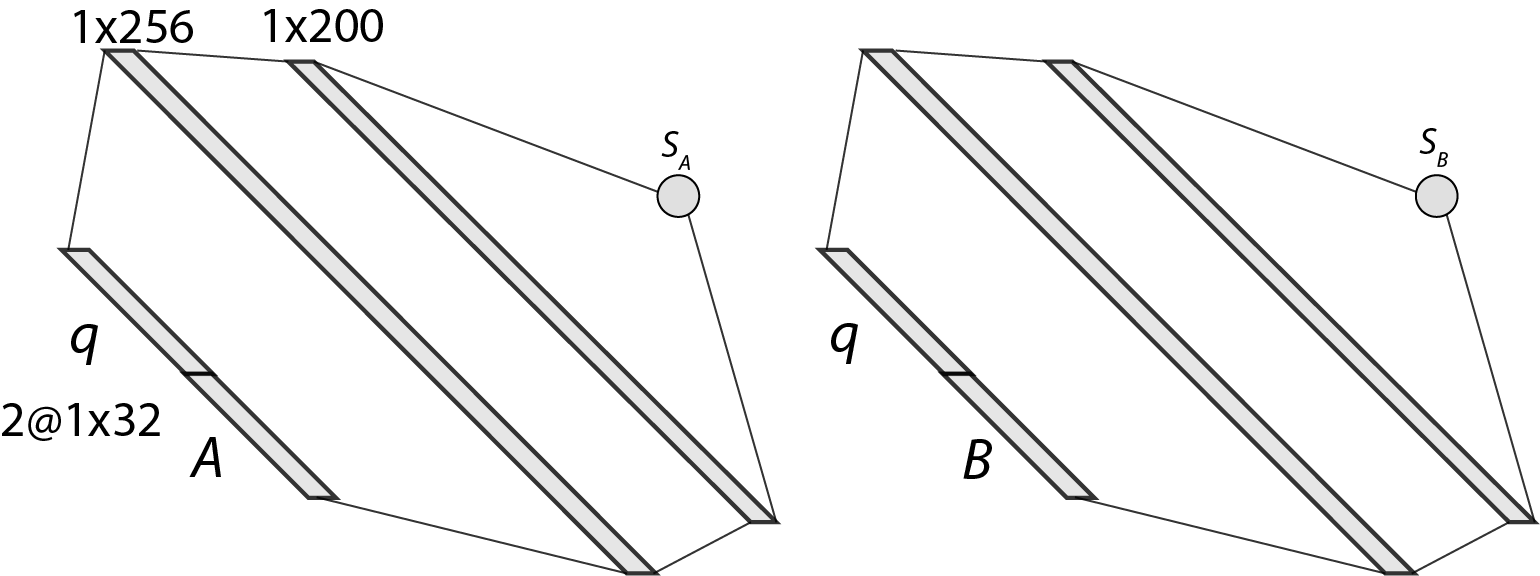}
\vspace{-6mm}
\caption{Similarity network architecture. Query vector $q$ and the vector representation of scan $A$ (for example) are concatenated and input to the network, which computes a similarity score $S_A$ between 0 and 1.}
\vspace{-6mm}
\label{fig:simnet}
\end{figure}

The architecture of our similarity network, shown in Fig \ref{fig:simnet}, uses three fully connected layers with ReLU activation functions which take as input two concatenated embedding vectors. The final layer outputs a scalar which is passed through a sigmoid activation function to produce a similarity score between 0 and 1.

\vspace{-1mm}
\subsection{Training}

To train the similarity network we construct a siamese network \cite{bromley1994signature}. We feed one sub-network with inputs $q$ and $s_A$, the other with inputs $q$ and $s_B$, and apply a form of margin ranking loss to the two resulting similarity scores. Because there is no \emph{a priori} ground truth for this task, either in terms of real-valued scores or relative-score labels, we construct labeled data from real data via several methods.

In one method, $q$ is a copy of $s_A$, a copy of $s_B$, or a combination of the two. Combinations are created by randomly generating a mask over a contiguous region of one input bitmap, say $s_A$, and then replacing the values in the masked area with values from the other bitmap, $s_B$. We call the masked area a subset selection. To avoid combinations which are equal parts $s_A$ and $s_B$, subset selections near one half the field of view are rejected.

In a second method, $s_A$ is copied twice and two different, random rotations are applied. The scans with the largest relative rotations become $s_A$ and $s_B$, while the middle scan becomes $q$. The loss function then prefers a higher similarity score for the scan whose rotation relative to $q$ is least. 

In a third method, a subset of a scan is randomly injected into either $s_A$ or $s_B$. $q$ represents the injected subset, while $s_A$ and $s_B$ are full scans, one of which contains $q$. A variant of this method is to rotate $q$ before insertion into $s_A$ or $s_B$.

During training, these methods are applied to inputs as they are fed into the network according to a curriculum. Margin ranking loss requires setting a margin, and we find empirically that a value of 0.01 works well. We use shuffled batches of 256 and update weights with Adam.

\section{Query Evaluation}

Because the neural network trained to compute similarity (or distance) does not learn a proper distance metric, there is no straightforward way to prune items from consideration during query evaluation. The naive solution is to do a linear pass through the entire database in an arbitrary order, computing the similarity between $q$ and every scan in the database and keeping track of the top k.

The proposed approach constructs a sparse multi-graph over the embedded vectors. Embedded vectors are nodes, and edge $e_{ij}$ exists if $||v_i - v_j|| \leq \epsilon$. We determine $\epsilon$ empirically, based on the local continuity of the embedded space. To estimate $\epsilon$, we let scan $s_i$ have an embedded representation $\vec{v_i}$ and let $\vec{\epsilon}$ be any vector with magnitude $\epsilon$. If decoding interpolations between $\vec{v_i}$ and $\vec{v_i} + \vec{\epsilon}$ yields smoothly changing recreations, for any choice of $\vec{\epsilon}$, then we say the embedded space is continuous in an $\epsilon$-ball around $\vec{v_i}$.

At query time, nodes are initially randomly selected for evaluation. After an initial sampling period $t$, the random selection is paused and the neighbors of top-scoring nodes are immediately evaluated. Neighbors continue to be expanded until the results stop ending up in the top k. Since there are no guarantees on the distance function, the query evaluator must visit every node eventually to ensure completeness. However, the proposed query evaluator, defined in Algorithm 1, generates higher quality top-k results faster than the arbitrary linear scan. 

  \setlength{\textfloatsep}{0pt}
  \begin{algorithm}[ht]
  \caption{\textsc{Laser2Vec Query Evaluator}}
  \begin{algorithmic}[1]
  \State $\textbf{Input:}$ Database $D$ as a graph $G$, query vector $q$
  \State $\textbf{Output:}$ $k$ most similar tuples in $D$ to $q$, $K$
  \State $K \gets \emptyset$
  \State $E \gets \emptyset$ \Comment{The expanded set}
  \ForAll{$t$ iterations}
  \State $v_i \gets$ \textsc{getNewRandNode}($G, K, E$)
  \State $s \gets$ \textsc{computeSimilarity}$(v_i, q)$
  \State $K \gets K \cup (v_i, s)$ 
  \EndFor
  \State $E \gets E \cup$ \textsc{notTopK($K$)}
  \State $K \gets$ \textsc{topK($K$)}
  \While{$\exists v_i \in K$ with unexpanded neighbors}
  \State $v_j \gets$ \textsc{getUnexpandedNeighbor($v_i$)}
  \State $s \gets$ \textsc{computeSimilarity}$(v_j, q)$
  \If{$s > $ \textsc{minVal($K$)}}
  \State $K \gets K \cup (v_j, q)$
  \State $K \gets K \setminus$ \textsc{minVal($K$)}
  \State $E \gets E \cup$ \textsc{minVal($K$)}
  \EndIf
  \EndWhile
  \If{$G \setminus (K \cup E) \not = \emptyset$}
  \State \textbf{go to} 4
  \EndIf
  \State \Return $K$
  \end{algorithmic}
  \label{alg:queryplan}
  \end{algorithm}

\section{Results}

We evaluate the Laser2Vec system in three capacities: accuracy, robustness to noise, and efficiency. The first two subsections present qualitative and quantitative results for queries where $q$ is a whole scan and a subset of a scan, respectively. The third subsection demonstrates Laser2Vec's invariance and stability to various sensor parameters. The last subsection shows the result quality versus time of the proposed query evaluator relative to the baseline linear scan. 

\subsection{Whole Scan Retrieval}

Because ground truth labels on real data do not exist, we can only show qualitative results regarding the system's ability to return specific scans for specific queries. However, one quantitative measure we do have is to play back all the log files and record how many times the robot visited topologically similar locations, such as T-junctions. We can then verify that the top-k results for a query with such a scan contain at least one scan from each such episode in the data. Indeed, this is the motivating use case, although this type of validation is only feasible for relatively small datasets.

Table~\ref{table:3} shows the recall scores of Laser2Vec and baseline methods for these instances. Successful recall is defined as at least one scan from the a given episode appearing in the top-k results. That is, if a robot travelled through two T-junctions, corresponding to scans 100-127 and 342-379, then a recall score of 1.0 means the query result contains at least one scan from the range 100-127 and one scan from the range 342-379. In our dataset, the robot visited T-junctions 10 times, lobbies 20 times, and segments of hallway 45 times. The value of $k$ for each query was 10 times the number of occurrences, so 100, 200, and 450, respectively. The entries in the table are the median values of 11 trials, each using unique query scans sampled from each location type.

The Raw Scan method computes the Euclidean distance between raw scans. FLIRT \cite{tipaldi2010flirt} and FALKO \cite{kallasi2016fast} are hand-engineered descriptors and are matched using RANSAC \cite{fischler1981random} and a nearest neighbor algorithm, respectively. FALKONet is a deep network we trained to produce similarity scores like the proposed method, but uses flattened FALKO descriptors instead of embedded representations.

Next, we present two experiments designed to test the system's ordinal consistency. If the system is shown to produce a reasonable result once (Table~\ref{table:3}), and also shown to be ordinally consistent under many arbitrary transformations, this is evidence that the system will produce reasonable results over a wide range of possible inputs. Here, we measure ordinal consistency with the Spearman rank correlation coefficient, or Spearman's $\rho$, which is the Pearson correlation coefficient calculated on the rank variables.

In the first experiment, multi-noise single-scan (MNSS), we take a single scan and populate a database by applying varying levels of noise along a single dimension. Specifically, we apply rotation. A top-k query is then run on the database and the ranking of the results is compared to the ranking of the magnitude of the applied noise. 

The second experiment, single-noise multi-scan (SNMS), applies identical transformations, in this case rotation by $\theta \sim \mathcal{U}(0, \pi / 4)$, to an existing set of scans. Spearman's $\rho$ is again calculated between the top-k results from the original database and the uniformly transformed database. 

Table \ref{table:molasses} shows the results for the MNSS and SNMS experiments. In SNMS we expect $\rho$ to vary less, since transforming the entire database should not perturb relative similarity assessments, while in MNSS, we expect $\rho$ to decrease slightly as $k$ increases, since the mapping between transformation and similarity may be non-trivial. Some of the baselines do well in SNMS, but are not as robust to noise and therefore do poorly in MNSS compared to Laser2Vec.

\begin{table}
\centering
\caption{Median episode recall scores}
\vspace{-1mm}
\normalsize
\begin{tabular}{ | c | c | c | c |}
\hline
  Location type: & T-junction & Lobby & Hallway \\
\hline
\hline
 FLIRT+RSC & 0.8 & 0.8 & 0.87 \\
\hline
 FALKO+NN & 0.9 & 0.8 & 0.91 \\
\hline
 Raw Scan & 0.3 & 0.35 & 0.38 \\
\hline
 FALKONet & 0.5 & 0.7 & 0.69 \\
\hline
 Laser2Vec & 0.9 & 0.9 & 0.96 \\
\hline
\end{tabular}
\label{table:3}
\end{table}

\begin{table}
\centering
\caption{Spearman's $\rho$ for the MNSS experiment (left) and the SNMS experiment (right), using whole scans. All entries are means of 10 queries.}
\vspace{-1mm}
\normalsize
\begin{tabular}{ | c | c | c | c || c | c | c |}
\hline
  Query size: & $50$ & $100$ & $200$ & $50$ & $100$ & $200$ \\
\hline
\hline
 FLIRT+RSC & 0.69 & 0.63 & 0.58 & 0.87 & 0.80 & 0.77 \\
\hline
 FALKO+NN & 0.71 & 0.68 & 0.59 & 0.89 & 0.86 & 0.81 \\
\hline
 Raw Scan & 0.45 & 0.39 & 0.27 & 1.0 & 1.0 & 1.0 \\
\hline
 FALKONet & 0.32 & 0.21 & 0.17 & 0.45 & 0.32 & 0.24 \\
\hline
 Laser2Vec & 0.84 & 0.78 & 0.69 & 0.76 & 0.71 & 0.68 \\
\hline
\end{tabular}
\label{table:molasses}
\end{table}

\subsection{Subset Scan Retrieval}

Unlike similarity for whole scans, we can construct datasets for subset scan retrieval which have ground truth labels. We do this with a method similar to that described in the neural network training sections. Ten templates, including items such as doorways, couches, humans, left turns, and convex corners are selected from raw scans and then injected into a subset of scans in the database. The database is then queried using the templates as input to generate $q$, and the results are analyzed for the percentage of the top-k scans which also contain the template. We conduct two recall experiments, both shown in Table \ref{table:big}. In the first experiment the subset scans are added to the database without any noise, and with the identical heading, and in the second experiment Gaussian noise and rotation are applied to the subset before insertion. In both cases, 500 templates were injected into a database with 10,000 vectors. $k$ was set to 200. In both experiments, Raw Scan similarity was computed by comparing only the depth readings for which the subset existed, since computing similarity over all possible moving window locations, essentially a 1D convolution, would be prohibitively expensive.

\begin{table*}[t]
\centering
\normalsize
\begin{tabular}{ | c | c | c | c | c | c | c | c | c | c | c |}
\hline
  Template type: & Left & Right & Couch & Human & Elevator & Hall & Two Doors & Fire Door & Conv C. & Conc C. \\
\hline
\hline
 FLIRT+RSC & 0.85 & 0.89 & 0.94 & 0.72 & 0.82 & 0.97 & 0.96 & 0.71 & 0.93 & 0.96 \\
\hline
 FALKO+NN & 0.88 & 0.80 & 0.96 & 0.75 & 0.90 & 0.95 & 0.96 & 0.76 & 0.97 & 0.98 \\
\hline
 Raw Scan & 1.0 & 1.0 & 1.0 & 1.0 & 1.0 & 1.0 & 1.0 & 1.0 & 1.0 & 1.0 \\
\hline
 FALKONet & 0.52 & 0.50 & 0.29 & 0.28 & 0.49 & 0.63 & 0.47 & 0.53 & 0.46 & 0.47 \\
\hline
 Laser2Vec & 0.71 & 0.70 & 0.73 & 0.51 & 0.61 & 0.75 & 0.56 & 0.59 & 0.72 & 0.74 \\
\hline
\hline
 FLIRT+RSC & 0.18 & 0.19 & 0.14 & 0.07 & 0.08 & 0.13 & 0.22 & 0.09 & 0.35 & 0.29 \\
\hline
 FALKO+NN & 0.21 & 0.17 & 0.19 & 0.09 & 0.11 & 0.16 & 0.22 & 0.11 & 0.33 & 0.34 \\
\hline
 Raw Scan & 0.06 & 0.07 & 0.05 & 0.05 & 0.06 & 0.04 & 0.08 & 0.05 & 0.05 & 0.07 \\
\hline
 FALKONet & 0.21 & 0.19 & 0.09 & 0.05 & 0.14 & 0.29 & 0.10 & 0.11 & 0.28 & 0.28 \\
\hline
 Laser2Vec & 0.61 & 0.62 & 0.69 & 0.31 & 0.52 & 0.72 & 0.42 & 0.47 & 0.68 & 0.67 \\
\hline
\end{tabular}
\caption{Subset selection accuracy for identical subsets (top) and noised subsets (bottom). Exact pattern matching works well without noise, but these methods do not produce reliable results on novel, but qualitatively similar, inputs. Laser2Vec accuracy also suffers when noise is added, especially for smaller templates, but overall generalizes better than other methods.}
\vspace{-7mm}
\label{table:big}
\end{table*}

\subsection{Invariance and Stability}

The representation of scans as 2D bitmaps naturally creates invariance to the field of view (FOV) of the laser, provided the embeddings are trained on data which has at least as large a FOV as the data at hand. The discretization provided by the bitmap also promotes stability with respect to laser angular resolution. Of course, drastically under-sampling can cause non-detection of informative features, and recovery from these scenarios is not guaranteed. However, if at least one observation registers in each pixel, laser resolution has no effect on the system's performance. For example, our 256x256 bitmap representation is designed to represent laser scans with a maximum range of 10 meters. This produces pixel side-lengths of 8cm, which for most modern depth sensors far exceeds the expected noise.

Since every scan is placed at the center of the image, translational invariance within the image is not a property we are concerned with. The variance due to a robot translating around the world and moving closer to and farther from objects is necessary for the system to differentiate between different scenarios. Similarly, rotational invariance is something we explicitly train against, since we want the network to distinguish between scenes which have been rotated.

\subsection{Query and Representation Efficiency}

To test the efficiency of the proposed query evaluator, we compare the quality of the partial result to the expected result of the naive method, as a function of the number of items evaluated. Specifically, we measure the normalized residual aggregate distance between the top $k$ after $t$ scans are evaluated, and the top $k$ after all scans are evaluated. Suppose the result after $t$ evaluations is $S_t$, and suppose the final result is $S_T$, where $|S_t| = |S_T| = k$. Further, let the function $\delta(q, v): \mathbb{R}^{32} \rightarrow (0, 1)$ represent the network which computes similarity scores. The residual aggregate distance in the top $k$ result after $t \in \{ k, \ldots, T \}$ evaluations is then

\vspace{-2mm}
\begin{equation*}
d_t = \sum_{v_i \in S_t}^k (1 - \delta(q, v_i)).
\end{equation*}
\vspace{-2mm}

\begin{figure}
\includegraphics[width=\columnwidth]{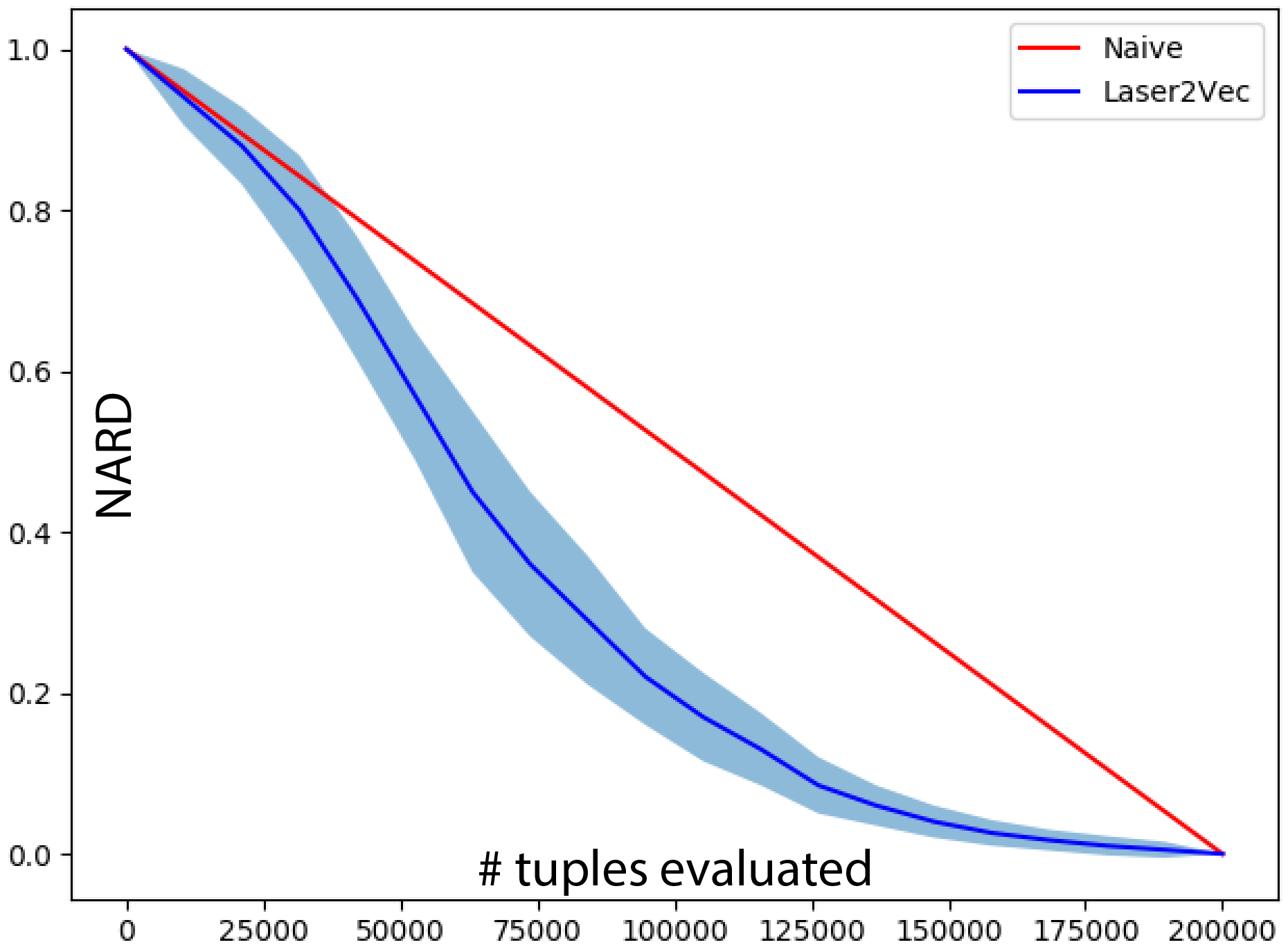}
\caption{Normalized aggregate residual distance (NARD) as a function of number of tuples evaluated. Mean (dark blue) and variance (light blue) for $N = 100$. Expected value for a naive pass in arbitrary order is shown in red.}
\label{fig:qe}
\end{figure}

Figure \ref{fig:qe} shows the mean and variance of the normalized residual aggregate distance, $\frac{d_t - d_T}{d_k - d_T}$, for 100 random queries. Here, $d_k$ is the aggregate residual distance after the first $k$ evaluations. Selecting tuples for evaluation at random, produces linear top-k quality improvement over time, in expectation. The optimal curve would drop to zero after evaluating just $k$ tuples. The reason we use a normalized residual is that some queries may have vastly different raw aggregate distance values since $q$ may be similar to many items in the database or very few, and this fundamentally limits the quality of a given top-k result for specific queries.

The memory efficiency of the Laser2Vec system compares favorably to the other methods. Raw scans, FLIRT descriptors, and FALKO descriptors require 4324, 7800, and 8384 bytes per scan, respectively. Our embedded representations require only 128. A laser running at 20Hz could run over 60 hours before occupying 1GB of memory using Laser2Vec.

\section{Discussion}

Laser2Vec is effective in terms of recall ability and efficiency per evaluation during query time. However, there are two main drawbacks, both of which stem from using neural networks as distance function approximators. First, the query times are comparatively slow - on the order of 1 sec per 100K items, since data must be sent from the database to the network. Possible optimizations include loading the weights into custom middleware written in a compiled language, or evaluating tuples concurrently. Additionally, meta data may be used to prune more items. For instance, each scan has a timestamp, and since the robot can only move so fast, in some cases evaluating both $s_{t}$ and $s_{t+1}$ may not be necessary, provided the distance function is robust enough.

The second drawback is that in order to guarantee optimal query results, every scan must be evaluated. However, practitioners probably do not need to know exactly which laser scan is the most similar to the query. Rather, they need to know roughly during which times signals resembling the query signal occurred in order investigate further. If the query result points them to one of the few dozen frames near the optimal solution, that is good enough. Allowing approximate results may make it possible to prune items more aggressively, further decreasing query times. 

\section{Conclusion}

\vspace{-1mm}

In this paper we presented Laser2Vec, a system for answering similarity-based top-k queries for robotic perception data. We demonstrated the query evaluator efficiency, accuracy, and robustness on real-world data, and introduced new experimental methods to evaluate such systems in the absence of ground truth. Future work includes the optimizations mentioned in $\S$8, as well as extensions to the representation of the data that allow users to annotate results with data more readily usable by relational database systems.

\vspace{-1mm}

\bibliography{laser2vec}

\begin{thebibliography}{10}

\bibitem{azizpour2015generic}
H.~Azizpour, A.~Sharif~Razavian, J.~Sullivan, A.~Maki, and S.~Carlsson.
\newblock From generic to specific deep representations for visual recognition.
\newblock In {\em Proceedings of the IEEE conference on computer vision and
  pattern recognition workshops}, pages 36--45, 2015.

\bibitem{babenko2014neural}
A.~Babenko, A.~Slesarev, A.~Chigorin, and V.~Lempitsky.
\newblock Neural codes for image retrieval.
\newblock In {\em European conference on computer vision}, pages 584--599.
  Springer, 2014.

\bibitem{biswas20161}
J.~Biswas and M.~Veloso.
\newblock The 1,000-km challenge: Insights and quantitative and qualitative
  results.
\newblock {\em IEEE Intelligent Systems}, 2016.

\bibitem{bromley1994signature}
J.~Bromley, I.~Guyon, Y.~LeCun, E.~S{\"a}ckinger, and R.~Shah.
\newblock Signature verification using a" siamese" time delay neural network.
\newblock In {\em Advances in neural information processing systems}, pages
  737--744, 1994.

\bibitem{deshpande2010efficient}
D.~P. P.~M. Deshpande.
\newblock Efficient rknn retrieval with arbitrary non-metric similarity
  measures.
\newblock 2010.

\bibitem{deshpande2008efficient}
P.~M. Deshpande, K.~Kummamuru, et~al.
\newblock Efficient online top-k retrieval with arbitrary similarity measures.
\newblock In {\em Proceedings of the 11th international conference on Extending
  database technology: Advances in database technology}, pages 356--367. ACM,
  2008.

\bibitem{fischler1981random}
M.~A. Fischler and R.~C. Bolles.
\newblock Random sample consensus: a paradigm for model fitting with
  applications to image analysis and automated cartography.
\newblock {\em Communications of the ACM}, 24(6), 1981.

\bibitem{hinton2006reducing}
G.~E. Hinton and R.~R. Salakhutdinov.
\newblock Reducing the dimensionality of data with neural networks.
\newblock {\em science}, 313(5786):504--507, 2006.

\bibitem{ilyas2008survey}
I.~F. Ilyas, G.~Beskales, and M.~A. Soliman.
\newblock A survey of top-k query processing techniques in relational database
  systems.
\newblock {\em ACM Computing Surveys (CSUR)}, 40(4):11, 2008.

\bibitem{jang2018grasp2vec}
E.~Jang, C.~Devin, V.~Vanhoucke, and S.~Levine.
\newblock Grasp2vec: Learning object representations from self-supervised
  grasping.
\newblock {\em arXiv preprint arXiv:1811.06964}, 2018.

\bibitem{jiang2017glass}
J.~Jiang, R.~Miyagusuku, A.~Yamashita, and H.~Asama.
\newblock Glass confidence maps building based on neural networks using laser
  range-finders for mobile robots.
\newblock In {\em 2017 IEEE/SICE International Symposium on System Integration
  (SII)}, pages 405--410. IEEE, 2017.

\bibitem{kallasi2016fast}
F.~Kallasi, D.~L. Rizzini, and S.~Caselli.
\newblock Fast keypoint features from laser scanner for robot localization and
  mapping.
\newblock {\em IEEE Robotics and Automation Letters}, 1(1):176--183, 2016.

\bibitem{kingma2014adam}
D.~P. Kingma and J.~Ba.
\newblock Adam: A method for stochastic optimization.
\newblock {\em arXiv preprint arXiv:1412.6980}, 2014.

\bibitem{kingma2013auto}
D.~P. Kingma and M.~Welling.
\newblock Auto-encoding variational bayes.
\newblock {\em arXiv preprint arXiv:1312.6114}, 2013.

\bibitem{krizhevsky2012imagenet}
A.~Krizhevsky, I.~Sutskever, and G.~E. Hinton.
\newblock Imagenet classification with deep convolutional neural networks.
\newblock In {\em Advances in neural information processing systems}, pages
  1097--1105, 2012.

\bibitem{lange2011efficient}
D.~Lange and F.~Naumann.
\newblock Efficient similarity search: arbitrary similarity measures, arbitrary
  composition.
\newblock In {\em Proceedings of the 20th ACM international conference on
  Information and knowledge management}, pages 1679--1688. ACM, 2011.

\bibitem{li2017deep}
J.~Li, H.~Zhan, B.~M. Chen, I.~Reid, and G.~H. Lee.
\newblock Deep learning for 2d scan matching and loop closure.
\newblock In {\em IROS}. IEEE, 2017.

\bibitem{mikolov2013distributed}
T.~Mikolov, I.~Sutskever, K.~Chen, G.~S. Corrado, and J.~Dean.
\newblock Distributed representations of words and phrases and their
  compositionality.
\newblock In {\em NIPS}, 2013.

\bibitem{moll2017exploring}
O.~Moll, A.~Zalewski, S.~Pillai, S.~Madden, M.~Stonebraker, and V.~Gadepally.
\newblock Exploring big volume sensor data with vroom.
\newblock {\em Proceedings of the VLDB Endowment}, 10(12):1973--1976, 2017.

\bibitem{nelson2016building}
P.~Nelson, C.~Linegar, and P.~Newman.
\newblock Building, curating, and querying large-scale data repositories for
  field robotics applications.
\newblock In {\em Field and Service Robotics}, pages 517--531. Springer, 2016.

\bibitem{nicolai2016deep}
A.~Nicolai, R.~Skeele, C.~Eriksen, and G.~A. Hollinger.
\newblock Deep learning for laser based odometry estimation.
\newblock In {\em RSS workshop Limits and Potentials of Deep Learning in
  Robotics}, 2016.

\bibitem{niemueller2012generic}
T.~Niemueller, G.~Lakemeyer, and S.~S. Srinivasa.
\newblock A generic robot database and its application in fault analysis and
  performance evaluation.
\newblock In {\em Intelligent Robots and Systems (IROS), 2012 IEEE/RSJ
  International Conference on}, pages 364--369. IEEE, 2012.

\bibitem{pfeiffer2017perception}
M.~Pfeiffer, M.~Schaeuble, J.~Nieto, R.~Siegwart, and C.~Cadena.
\newblock From perception to decision: A data-driven approach to end-to-end
  motion planning for autonomous ground robots.
\newblock In {\em ICRA}. IEEE, 2017.

\bibitem{razavian2016visual}
A.~S. Razavian, J.~Sullivan, S.~Carlsson, and A.~Maki.
\newblock Visual instance retrieval with deep convolutional networks.
\newblock {\em ITE Transactions on Media Technology and Applications},
  4(3):251--258, 2016.

\bibitem{saxena2014robobrain}
A.~Saxena, A.~Jain, O.~Sener, A.~Jami, D.~K. Misra, and H.~S. Koppula.
\newblock Robobrain: Large-scale knowledge engine for robots.
\newblock {\em arXiv preprint arXiv:1412.0691}, 2014.

\bibitem{szegedy2015going}
C.~Szegedy, W.~Liu, Y.~Jia, P.~Sermanet, S.~Reed, D.~Anguelov, D.~Erhan,
  V.~Vanhoucke, and A.~Rabinovich.
\newblock Going deeper with convolutions.
\newblock In {\em CVPR}, pages 1--9. IEEE, 2015.

\bibitem{thomas2014rosbag}
D.~Thomas, T.~Field, J.~Leibs, and J.~Bowman.
\newblock Rosbag-ros wiki.
\newblock {\em URL: http://wiki. ros. org/rosbag}, 2014.

\bibitem{tipaldi2010flirt}
G.~D. Tipaldi and K.~O. Arras.
\newblock Flirt-interest regions for 2d range data.
\newblock In {\em ICRA}, pages 3616--3622. IEEE, 2010.

\bibitem{wilkinson2015efficient}
E.~Wilkinson and T.~Takahashi.
\newblock Efficient aspect object models using pre-trained convolutional neural
  networks.
\newblock In {\em International Conference on Humanoid Robots}, pages 284--289.
  IEEE, 2015.

\bibitem{xin2007progressive}
D.~Xin, J.~Han, and K.~C. Chang.
\newblock Progressive and selective merge: computing top-k with ad-hoc ranking
  functions.
\newblock In {\em Proceedings of the 2007 ACM SIGMOD international conference
  on Management of data}, pages 103--114. ACM, 2007.

\bibitem{yu2016top}
A.~Yu, P.~K. Agarwal, and J.~Yang.
\newblock Top-$ k $ preferences in high dimensions.
\newblock {\em IEEE Transactions on Knowledge and Data Engineering},
  28(2):311--325, 2016.

\bibitem{zhu2016evaluating}
L.~Zhu, F.~Liu, W.~Meng, Q.~Ma, Y.~Wang, and F.~Yuan.
\newblock Evaluating top-n queries in n-dimensional normed spaces.
\newblock {\em Information Sciences}, 374:255--275, 2016.

\end{thebibliography}

\end{document}